\pdfoutput=1
\documentclass[11pt]{article}
\usepackage{enumitem}
\usepackage[final]{acl}
\usepackage{booktabs} 
\usepackage{times}
\usepackage{latexsym}

\usepackage[T1]{fontenc}
\usepackage[utf8]{inputenc}
\usepackage{tcolorbox}
\usepackage{microtype}

\usepackage{inconsolata}

\usepackage{graphicx}

\usepackage{colortbl}
\usepackage{pifont}
\usepackage{amssymb}
\usepackage{amsmath}% http://ctan.org/pkg/amsmath

\usepackage{makecell} % If not already included, this is for `\makecell`

\usepackage{subfigure}
\newcommand{\ours}{\textsc{RAG-GUI}}

\definecolor{lightcoral}{HTML}{F08080}
\definecolor{green}{HTML}{009B55}
\title{Retrieval-augmented GUI Agents with Generative Guidelines}

\author{Ran Xu$^1$, Kaixin Ma$^2$, Wenhao Yu$^2$, Hongming Zhang$^2$, \\
\bf Joyce C. Ho$^1$, Carl Yang$^1$, Dong Yu$^2$
 \\
  $^1$ Emory University \quad  $^2$ Tencent AI Lab \\
  \texttt{ran.xu@emory.edu}}

\begin{document}
\maketitle
\begin{abstract}
GUI agents powered by vision-language models (VLMs) show promise in automating complex digital tasks.
% but struggle in real-world settings due to limited training data and task complexity. 
% requires knowledge beyond the knowledge in the LLMs' parameters either because of the software updating or rare scenarios.
However, their effectiveness in real-world applications is often limited by scarce training data and the inherent complexity of these tasks, which frequently require long-tailed knowledge covering rare, unseen scenarios.
% While prior work uses web tutorials to synthesize training data, these static datasets limit adaptability. 
We propose \ours{}
% \footnote{Short for \textbf{T}utorial-\textbf{A}ugmented Agents with \textbf{P}lug-and-Play \textbf{T}ask-\textbf{A}daptive \textbf{G}uideline Generation.}
, a lightweight VLM that leverages web tutorials at inference time. 
% Acting as an adapter between the agent and tutorials, 
\ours{} is first warm-started via supervised finetuning (SFT) and further refined through self-guided rejection sampling finetuning (RSF).
Designed to be model-agnostic, \ours{} functions as a generic plug-in that enhances any VLM-based agent.
Evaluated across three distinct tasks, it consistently outperforms baseline agents and surpasses other inference baselines by 2.6\% to 13.3\% across two model sizes, demonstrating strong generalization and practical plug-and-play capabilities in real-world scenarios.
\end{abstract}

\section{Introduction}
Graphical User Interface (GUI) agents have emerged as powerful tools capable of automating complex interactions across diverse digital platforms, including web browsers~\citep{zhou2024webarena,he-etal-2024-webvoyager}, computer use \citep{xie2024osworld}, and mobile applications~\citep{rawles2023androidinthewild,rawles2024androidworld}.
% Graphical User Interfaces (GUIs) serve as the primary interface through which users interact with software, from web applications to complex enterprise systems~\citep{cheng-etal-2024-seeclick,}. 
Recent advances in vision-language models (VLMs) \citep{llava,bai2025qwen2} have greatly enhanced agents' abilities in grounding, visual context understanding, and reasoning, leading to notable progress in GUI-based interaction. However, these agents still struggle with real-world tasks due to their inherently complex, multi-step nature and the limited availability of high-quality training data~\citep{aksitov2023rest,sun2024genesis}.

To address these challenges, several studies leverage web tutorials, which provide step-by-step instructions and rich contextual information~\citep{xu2025agenttrek,ou2024synatra,zhang2025tongui}, to synthesize agent trajectories for model finetuning. However, the quality of such synthetic data remains variable, limiting flexibility and generalization to novel tasks.

We propose leveraging web tutorials as a non-parametric knowledge base at inference time to enhance agents’ adaptability across diverse tasks. While this setup resembles the retrieval-augmented generation (RAG) paradigm, it differs from standard RAG pipelines that rely on cleaned, chunked Wikipedia passages for QA~\citep{lewis2020retrieval,liu2025roserag,shi-etal-2024-replug}
and introduces two intrinsic challenges: (1) tutorials often encode procedural knowledge that is lost with fixed-length chunking, while leaving them unprocessed leads to long, noisy inputs that may degrade LLM performance~\citep{yu-etal-2024-chain}; and (2) tutorial relevance is not guaranteed—unlike QA corpora, tutorials may not align with the agent's task. 
Addressing these challenges is critical to fully leveraging tutorials for guiding agents towards desired behaviors.

Motivated by the above challenges, we introduce \ours{},
% \footnote{Short for \textbf{T}utorial-\textbf{A}ugmented \textbf{P}lug-and-play \textbf{A}gents with \textbf{S}ummarization}
a lightweight VLM as an \emph{adapter} between the agent backbone and the tutorial during inference. 
\ours{} is designed to: (1) assess the relevance between the current task (query and prior actions) and a given tutorial; and (2) generate a useful guidance from the relevant tutorial to assist task completion.
Training \ours{} begins with supervised finetuning (SFT) with synthetic relevance labels generated by a teacher VLM to bootstrap learning. We then use  rejection sampling that allows \ours{} to refine both relevance prediction and summarization, based on the assumption that \emph{better guidance improves agent performance}. 
\ours{} is model-agnostic and can serve as a plug-and-play module for any VLM-based agent, making it broadly applicable across different  tasks.

% We evaluate \ours{} on three tasks with two backbones, and find that it consistently improves the performance of VLM-based agents on the inference stage and often perform better than agent trained on synthetic trajecties collected from tutorials. 
% Notably, on the online evaluation benchmark Android world, \ours{} leads to 13.3\% and 10.7\% absolute gains for improving over 7B/72B backbones.
% Furthermore, \ours{} narrows the gap with training-based methods and even surpasses some of them in online evaluations, highlighting its strong generalization across diverse, real-world scenarios.

We evaluate \ours{} on three tasks with two backbones, consistently improving VLM-based agents at inference and often outperforming agents trained on synthetic tutorial trajectories. On the online AndroidWorld benchmark, \ours{} achieves 13.3\% and 10.7\% absolute gains for 7B and 72B backbones, respectively. Moreover, it narrows or surpasses the gap with training-based methods in real-world settings, highlighting its strong generalization across diverse, real-world scenarios.

% of a lightweight white-box LLM that
% functions as a controller
% In this paper, we propose to leverage web tutorials as a non-parametric knowledge base at inference time, thus significantly expanding agents' adaptability to diverse tasks. 
% Our setting mimic the retrieval-augmented generation (RAG) setting, yet different from the RAG pipeline in general domain where cleaned, chunked Wikipedia corpus are provided for question asnwering \citep{lewis2020retrieval,petroni-etal-2021-kilt,shi-etal-2024-replug}, leveraging tutorial presents several intrinsic challenges: (1) tutorial ofen contains precedual knwpledge, direct thinking will lose such information, while do not process yield long and noisy text
% (2) irrelevance: different than knowledge based QA task, there is no guarantee that there are relevant tutorial on the web. Therefore, it is necessary to design  tailored techniques to adapt retrieval to LLMs.
% However, different from existing 
% Our approach involves first retrieving relevant web tutorials using a small, task-specific Vision Language Model (LLM) trained explicitly to assess the relevance of tutorial content and provide succinct summaries tailored to the agent's current context. 

\section{Related Works}
Enabling large language models (LLMs) and vision-language models (VLMs) to function as capable GUI agents has been a vibrant area of research. Early approaches~\citep{deng2023mind2web,gur2023real} improved web agents by leveraging raw HTML elements but did not incorporate visual information. More recent works have demonstrated the promise of VLMs for GUI tasks, capitalizing on their strong visual understanding~\citep{zheng2024gptvision,yan2023gpt,hong2024cogagent}. Concurrently, efforts such as~\citep{gou2025navigating,wu2025osatlas,xu2024aguvis} focus on collecting large-scale training data to enhance model capabilities.

Closest to our approach, AgentTrek~\citep{xu2025agenttrek} and TongUI~\citep{zhang2025tongui} leverage web tutorials by synthesizing training trajectories to improve agent performance. Our work takes a complementary perspective: rather than relying solely on synthetic data generation, we directly integrate tutorial-based guidance at inference time through a lightweight, adaptive retrieval-augmented framework. This design enables flexible, plug-and-play enhancement of VLM-based GUI agents without the need for extensive retraining.

Retrieval-augmented generation (RAG) serves as a powerful technique for knowledge intensive tasks under both text-only~\citep{lewis2020retrieval,shi-etal-2024-replug,yu-etal-2024-chain,yu2024rankrag,xu-etal-2025-simrag} and multimodal scenarios~\citep{yu2025visrag}. However, in our scenario, leveraging tutorials poses unique challenges due to their length and the presence of potentially irrelevant or noisy context, which motivates our design of task-aware guidance generation. 

\section{Methodology}
\subsection{Preliminaries}
We formulate GUI tasks as a sequential decision-making problem and adopt the SeeAct framework~\citep{zheng2024gptvision}. Given a website state $s$, a task description $g \in \mathcal{G}$, and an action space $\mathcal{A}$, the agent generates a sequence of actions $A = (a_1, a_2, \ldots, a_n) \in \mathcal{A}^n$ to complete the task. At $t$-th step, the VLM-based agent $\pi$ selects the next action $a_t$ based on the current environment observation $s_t$, the task description $g$, and the history of previous actions $A_t = (a_1, \ldots, a_{t-1})$ as 
$a_t = \pi(g, s_t, A_t)$. 
In our tutorial-augmented setting, we first retrieve a set of potentially relevant tutorials $\tau = \{\tau_1, \ldots, \tau_k\}$ using the task description $g$. A lightweight VLM $f_\theta$ then processes each tutorial to produce task-aware guidance as:
\begin{equation}
\nonumber
\setlength{\abovedisplayskip}{3pt}
\setlength{\belowdisplayskip}{3pt}
\hat{\tau}_{i,t} = f_\theta(g, s_t, A_t, \tau_i).
\end{equation}
Here, the guidance $\hat{\tau} = (\ell, \sigma)$ contains a binary relevance label $\ell$ with the task-aware guidance $\sigma$. 
The agent then generates the next action based on the current state, task description, action history, and the set of summaries identified as relevant:
\begin{equation}
\setlength{\abovedisplayskip}{4pt}
\setlength{\belowdisplayskip}{4pt}
\nonumber
\hat{a}_t = \pi(g, s_t, A_t, \hat{\sigma_t}),
\end{equation}
where   $\hat{\sigma}_t = \{\hat{\sigma}_{i,t} \mid 1\leq i \leq k, \ell_i=1 \}$ is the filtered set of guidance summarized from relevant tutorials. 
We emphasize that the agent policy $\pi$ is fixed without parameter update and and only the VLM $f_\theta$ is updated.

\subsection{Tutorial Collection}
Our initial phase involves constructing a comprehensive dataset of GUI tutorials from diverse open-source online repositories. We define a GUI tutorial as a resource that provides sequential, step-by-step instructions for executing a task within a GUI, incorporating both textual descriptions and visual screenshots. Following~\citet{qin2025ui}, we select two large-scale, image-text interleaved datasets, MINT~\citep{awadalla2024mint} and OmniCorpus~\citep{li2024omnicorpus}, as primary sources for potential tutorial data. Additionally, we incorporate articles crawled from the WikiHow website\footnote{\url{www.wikihow.com}} as a supplementary corpus, given its focus on how-to content.
Recognizing that MINT and OmniCorpus include a wide spectrum of topics beyond GUI instruction, we conduct a multi-stage filtering process to extract relevant high-quality tutorials. This process comprises three key steps: (1) FastText Filtering, (2) Deduplication, and (3) LLM Labeling.
The details for these steps are deferred to Appendix \ref{app:tutorial}. 
After all steps, we obtain a high-quality set of approximately 2.6M GUI tutorials from MINT, OmniCorpus and GUI tutorials, which serves as the task-adaptive guidance resource for the subsequent experiments.

\subsection{Optimization for Task-aware Guidance Generation $f_{\theta}$}
Using a frozen VLM to generate task-specific guidance often leads to suboptimal results, as these models are typically fine-tuned for general-purpose tasks. 
This can create a mismatch between what the adapter $f_{\theta}$ perceives as `good' guidance and what the agent $\pi$ effectively utilizes. 
To this end, we finetune $f_{\theta}$ to align the  generated guidance towards target tasks by leveraging the training data containing $\mathcal{D} = \{(g_i, s_i, A_i, a_i)\}_{i=1}^{\mathcal{|D|}}$ where $a_i$ is the ground-truth action step. 

\paragraph{SFT Warmup.} 
To warm up $f_{\theta}$ for the target application, we first learn an initial policy via supervised finetuning by imitating the behavior of an expert VLM $u$. Specifically, we use a frontier model, \texttt{GPT-4.1-mini}, to generate high-quality guidance on open-sourced datasets of (state, tutorial, action) pairs as $\mathcal{D}_{\text{SFT}}=\{(x, h)\}$, where $x = (g, s, A, \tau_i)$ is the input containing both environment states, actions and retrieve tutorial,  $h\sim u(x)$ is sampled from the expert VLM. 
This dataset is used to perform supervised fine-tuning on $f_{\theta}$ as $\mathcal{L}_{\text{SFT}}=-\mathbb{E}_{(x, h) \sim \mathcal{D}_{\text{SFT}}} \sum_{l=1}^{|h|} \log f_\theta\left(h_l \mid h_{<l}, x\right)$, providing a warm-up for downstream optimization.

\paragraph{Self-guided Rejection Sampling Finetuning.}  
Generating large-scale high-quality distillation data is often infeasible in real-world settings. To improve the model's ability to produce effective guidance from tutorials, we adopt the hypothesis that \emph{good guidance should help the agent take the correct action}. To that end, we repurpose the existing training tuples $(g, s_t, a_t)$ to first retrieve relevant tutorials $\tau$, then train the guidance generation model such that its output increases the likelihood of the agent selecting the correct action as (note that $\hat{\tau}_i = (\hat{\ell}_i, \hat{\sigma}_i)$):

\vspace{-2ex}
\begin{small}
\begin{align}
\setlength{\abovedisplayskip}{4pt}
\setlength{\belowdisplayskip}{4pt}
\nonumber
p(\hat{a}_t &= a_t | g, s_t, \tau_i) = \log \sum_{\hat{\tau}_i}p_{\theta}(\hat{\tau}_i|\tau_i)p(\hat{a}_t=a_t | g, s_t, \hat{\sigma}_i) \\
&\geq \underbrace{\mathbb{E}_{q} [\log p(\hat{a}_t=a_t | g, s_t, \hat{\sigma}_i)] + \mathbb{D}_{\text{KL}} (q~||~p_{\theta}(\hat{\tau_i}|\tau_i))}_{\text{ELBO}~~\mathcal{L}(p_{\theta}, q)}
\nonumber
\end{align}
% \end{equation}
\end{small}
% \noindent Instead of directly optimizing this probability, we instead optimize its ELBO $\mathcal{L}(p_{\theta}, q)$, where $q(\hat{\tau})$ is a variational posterior over the sampled guidance $\hat{\tau}$. The optimal $q$ is given by:
\noindent Rather than directly optimizing the marginal log-likelihood, we maximize its evidence lower bound (ELBO) $\mathcal{L}(p_{\theta}, q)$, where $q(\hat{\tau})$ is the posterior over the sampled guidance $\hat{\tau}$. The optimal $q$ is given by:

\vspace{-2ex}
\begin{small}
\begin{align}
\setlength{\abovedisplayskip}{2pt}
\setlength{\belowdisplayskip}{2pt}
\nonumber
q\left(\hat{\tau}_i \mid g, s_t, a_t, \tau_i\right) &=\frac{p_{\theta}\left(\hat{\tau}_i \mid \tau_i\right) p\left(\hat{a}_t=a_t \mid g, s_t, \hat{\sigma}_i\right)}{\sum_{\hat{\tau}^{\prime}} p_{\theta}\left(\hat{\tau}_i^{\prime} \mid \tau_i\right) p\left(\hat{a}_t=a_t \mid g, s_t, \hat{\sigma}_i^{\prime}\right)} \\ 
&\propto p_{\theta}(\hat{\tau}_i \mid \tau_i) \cdot p(\hat{a}_t=a_t \mid g, s_t, \hat{\sigma}_i).
\nonumber
\end{align}
\end{small}
% Assume the success probability $p(a_t \mid g, s_t, \hat{\tau}_i) \propto \mathbb{I}(\hat{a}_t=a_t)$, then we can derive the following rules for model sampling: 
Assume $p(a_t \mid g, s_t, \hat{\tau}) \propto \mathbb{I}[\hat{a_t} = a_t]$, i.e., the probability is nonzero only when the generated action matches the ground-truth. Then, the overall rejection sampling procedure is as follows:

\begin{enumerate}[leftmargin=0.45cm]
\setlength{\abovedisplayskip}{3pt}
\setlength{\belowdisplayskip}{3pt}
  \item \textbf{Sample guidance:}  For each tutorial $\tau_i$, draw $m$ candidate guidances
    \(
      \{\hat\tau_i^{(j)}\}_{j=1}^m \;\sim\; p_\theta(\,\hat\tau\mid\tau_i).
    \)
  \item \textbf{Guideline Filtering:}  For each Guideline $\hat\tau_i^{(j)}$, let
    \(
      \hat{a}_t^{(j)} = f(g,\,s_t,\,\hat\sigma_i^{(j)})
    \), then  only those $(\tau_i,\hat\tau_i^{(j)})$ for which
    $\hat{a}_t^{(j)}=a_t$ are kept, yielding a filtered set
    $\mathcal{D}_{\text{RSF}}$.
  \item \textbf{Fine-tune:}  Update $f_\theta$ by minimizing the RSF Loss of the retained guideline:
   \(\mathcal{L}_{\text{RSF}} = 
      \min
      \left(-\sum_{(\tau_i,\hat\tau)\in\mathcal{D}_{\text{RSF}}}
      \log p_\theta(\hat\tau\mid\tau_i)\right).
    \)
\end{enumerate}
In practice, if the model assigns conflicting relevance labels to the same tutorial across different generations—yet both instances result in the correct action—we discard these examples to prevent introducing ambiguity into the training of the guideline generator $f_{\theta}$.
\subsection{Model Inference}
At the inference stage, for example $(g, s_t, A_t, a_t)$ and retrieved tutorials $\tau=\{\tau_1, ..., \tau_k\}$, 
we first use the guideline generation model $f_{\theta}$ to generate relevant guidelines $(\ell_i, \sigma_i) = f_\theta(g, s_t, A_t, \tau_i)$ for each tutorial $\tau_i$. 
Next, we filter the generated guidelines to retain only those marked relevant and augment the original prompt with this filtered set as $\hat{\sigma}_t = \{{\sigma}_{i} \mid 1\leq i \leq k, \ell_i=1 \}$. 
The agent $\pi$ uses this augmented guideline to make its final action prediction as 
$
\hat{a}_t=\pi\left(g, s_t, A_t, \hat{\sigma}_t\right), 
$
which updates the environment state $s_{t+1}$ and action history $A_{t+1}$ for the subsequent step.

\begin{table*}[t]
\centering
\renewcommand\arraystretch{0.93}
\caption{The performance of \ours{} and baselines on three benchmarks. M2W, AC and AW stands for mm-Mind2Web, AndroidControl and AndroidWorld, respectively.}
\label{tab:main}
\resizebox{\linewidth}{!}{
\begin{tabular}{cc|c|ccc|ccc|ccc|cc}
\toprule
\multicolumn{2}{c|}{\textbf{Models}} & \textbf{AW} & \multicolumn{3}{c|}{\textbf{M2W - Cross Task}} & \multicolumn{3}{c|}{\textbf{M2W - Cross Website}} & \multicolumn{3}{c|}{\textbf{M2W - Cross Domain}} & \textbf{AC - High} &  \textbf{AC - Low} \\
\midrule
\bf Planner  & \bf Grounder  & \bf SR & \bf Ele. Acc & \bf Op. F1 & \bf Step SR & \bf Ele. Acc & \bf Op. F1 & \bf Step SR & \bf Ele. Acc & \bf Op. F1 & \bf Step SR & \bf Step Acc. & \bf Step Acc.\\
\midrule
% GPT-4 & SeeClick & 29.7 & -- & -- & 28.5 & -- & -- & 30.7 & -- & -- \\
% GPT-4  &  UGround & 45.1 & -- & -- & 44.7 & -- & -- & 44.6 & -- & -- \\
GPT-4o~\shortcite{hurst2024gpt}  &  SeeClick~\shortcite{cheng2024seeclick} & -- & 32.1 & -- & -- & 33.1 & -- & -- & 33.5 & -- & -- & 41.8 & 52.8 \\
GPT-4o~\shortcite{hurst2024gpt}  & UGround~\shortcite{gou2025navigating} & 32.8 & 47.7 & -- & -- & 46.0 & -- & -- & 46.6 & -- & -- & 48.4 & 62.4 \\
GPT-4V~\shortcite{achiam2023gpt} & OmniParse~\shortcite{lu2024omniparser} & -- &  42.4 & \textbf{87.6} & 39.4 & 41.0 & \textbf{84.8} & 36.5 & 45.5 & 85.7 & 42.0 & -- & -- \\
\multicolumn{2}{c|}{GPT-4o~\shortcite{hurst2024gpt}} & 34.5 (SOM) &  5.7 & 77.2 & 4.3 & 5.7 & 79 & 3.9 & 5.5 & \textbf{86.4} & 4.5 & 20.8 & 19.4 \\
\multicolumn{2}{c|}{Claude Computer Use~\shortcite{claude}} & 27.9 & 62.7 & 84.7 & 53.5 & 59.5 & 79.6 & 47.7 & 64.5 & 85.4 & 56.4 & 12.5 & 19.4 \\
\multicolumn{2}{c|}{AgentTrek 7B~\shortcite{xu2025agenttrek}} & -- & 45.5 & 84.9 & 40.9 & 40.8 & 82.8 & 35.1 & 48.6 & 84.1 & 42.1 & -- & -- \\
\midrule
\multicolumn{2}{l|}{Qwen2.5-VL-7B~\shortcite{bai2025qwen2}} & 22.0 & 57.9 & 82.5 & 45.3 & 58.6 & 80.0 & 41.6 & 57.5 & 83.4 & 45.2 & 52.9 & 72.5 \\
\multicolumn{2}{l|}{~~~~~~~~~~~~~~~~~~ w/ vanilla RAG} & 22.4 & 59.0 & 83.1 & 46.0 & 57.4 & 80.4 & 41.3 & 57.9 & 83.5 & 45.3 & 55.8 & 74.3 \\
\rowcolor{lightcoral!20}\multicolumn{2}{l|}{\ours{}-7B} & 35.3 & 63.9 & 84.1 & 51.5 & 60.7 & 83.6 & 46.1 & 60.3 & 84.5 & 47.8 & 59.6 & 81.6 \\
\multicolumn{2}{l|}{~~~~~~~~~~~~~~~~~~ w/o RSF} & 32.8 & 59.9 & 83.5 & 46.8 & 58.6 & 82.3 & 44.2 & 59.5 & 84.5 & 47.3 & 54.9 & 79.4 \\
\midrule
\multicolumn{2}{l|}{Qwen2.5-VL-72B~\shortcite{bai2025qwen2}} & 35.0 & 63.4 & 81.7 & 51.8 & 64.0 & 81.8 & 49.6 & 56.5 & 84.1 & 46.2 & 57.0 & 78.6 \\
\multicolumn{2}{l|}{~~~~~~~~~~~~~~~~~~ w/ vanilla RAG} & 37.5 & 58.6 & 82.9 & 45.8 & 63.8 & 81.8 & 49.0 & 60.0 & 81.6 & 49.8 & 56.2 & 76.7 \\
\rowcolor{teal!12}\multicolumn{2}{l|}{\ours{}-72B} & {45.7} & \textbf{69.5} & 85.0 & \textbf{56.8} & \textbf{68.1} & 83.1 & \textbf{53.0} & \textbf{66.3} & 85.9 & \textbf{55.0} & \textbf{60.7} & \textbf{84.6} \\
\multicolumn{2}{l|}{~~~~~~~~~~~~~~~~~~ w/o RSF} & 44.4 & 66.3 & 84.2 & 53.8 & 66.9 & 82.9 & 52.1 & 63.6 & 84.7 & 52.2 & 58.7 & 80.9 \\
% \textbf{Metric} &  EM & EM / Acc. & EM / Acc. & EM / F1 & EM / F1 & Acc.  \\
\midrule
\multicolumn{3}{c|}{---} & \multicolumn{11}{c}{\textit{Training-based LLMs using in-distribution training data. For Reference Only.}} \\
\midrule
\multicolumn{2}{c|}{AgentTrek 7B (w/ M2W)~\shortcite{xu2025agenttrek}} & -- & 60.8 & 88.9 & 55.7 & 57.6 & 88.1 & 51.4 & 56.0 & 87.5 & 52.6 & -- & --\\
% \multicolumn{2}{c|}{UI-TARS 2B} & 62.3 & 90.0 & 56.3 & 58.5 & 87.3 & 50.8 & 58.8 & 89.6 & 52.3 \\
\multicolumn{2}{c|}{Aguvis 7B~\shortcite{xu2024aguvis}} & -- & 64.2 & 89.8 & 60.4 & 60.7 & 88.1 & 54.6 & 60.4 & 89.2 & 56.6 & 61.5 & 80.5 \\
\multicolumn{2}{c|}{Aguvis 72B~\shortcite{xu2024aguvis}} & 26.1 & 69.5 & 90.8 & 64.0 & 62.6 & 88.6 & 56.5 & 63.5 & 88.5 & 58.2 & 66.4 & 84.4 \\
\multicolumn{2}{c|}{UI-TARS 7B~\shortcite{qin2025ui}} & 33.0 & 73.1 & 92.2 & 67.1 & 68.2 & 90.9 & 61.7 & 66.6 & 90.9 & 60.5 & 72.5 & 90.8 \\
\multicolumn{2}{c|}{UI-TARS 72B~\shortcite{qin2025ui}} & \textbf{46.6} & 74.7 & 92.5 & 68.6 & 72.4 & 91.2 & 63.5 & 68.9 & 91.8  &  62.1 & 74.7 & 91.3\\
\bottomrule
\end{tabular}
}
\end{table*}

\section{Experiments}
\subsection{Experiment Setups}
\paragraph{Datasets and Evaluation Metrics} We evaluate \ours{} on one challenging out-of-distribution online dataset Android World~\citep{rawles2024androidworld} and two in-distribution offline datasets Android Control~\citep{li2024effects} and Multimodal-Mind2Web~\citep{deng2023mind2web}. The details for these datasets are in Appendix \ref{app:datasets}.

\paragraph{Baselines}
We compare against the following baselines:
(1) \textbf{Inference-based methods without tutorials}, including Seeclick~\citep{cheng-etal-2024-seeclick}, UGround~\citep{gou2025navigating}, OmniParse~\citep{lu2024omniparser}, GPT-4o~\citep{hurst2024gpt}, and Claude~\citep{claude};
(2) \textbf{Tutorial-based inference method}: vanilla RAG~\citep{lewis2020retrieval}, which augments tasks, states, and previous actions with top retrieved tutorials;
(3) \textbf{Tutorial-based synthetic data generation}: AgentTrek~\citep{xu2025agenttrek}, which finetunes the VLM using synthetic trajectories generated from web tutorials. 
We also report results from finetuning-based methods that leverage large amounts of grounding data and agent trajectories~\citep{xu2024aguvis,qin2025ui}, but mainly for reference. Designing approaches to incorporate our tutorial guideline generation into finetuning is an interesting direction for future work.

\paragraph{Implementation Details} Our guideline generation model $f_\theta$ is built upon the \texttt{Qwen-2.5-VL-7B} backbone model. To create task-specific guideline for training, we utilize AitW~\citep{rawles2023androidinthewild}, AMEX~\citep{chai2024amex}, and GUIAct (web-multi and web-single)~\citep{chen2024guicourse} as seed datasets. 
We use E5~\citep{wang2022text} as the default embedding model for tutorial retrieval.
For the SFT warmup, we train the backbone model for 1 epoch using a learning rate of $1e-5$ and a cosine scheduler. During rejection sampling, we set the temperature to 1.0. For the RSF stage, we continue training the model for one additional epoch, building on the SFT checkpoint, with a learning rate of $5e-6$. For evaluation, we use \texttt{Qwen-2.5-VL-7B/72B}~\citep{bai2025qwen2} as the agent backbone model.

\subsection{Main Experiments}
Table \ref{tab:main} exhibits the experiment results on three tasks. We have the following findings:
\paragraph{Experiment on Offline Tasks} Evaluations on AndroidControl and mm-Mind2Web reveal that naively incorporating tutorials via a standard RAG pipeline yields limited improvements, particularly for smaller backbones (7B), due to LLMs’ constrained ability to process tutorials directly. In contrast, \ours{} achieves notable gains (4.4\% on Mind2Web and 6.3\% on AndroidControl) and narrow the gap between training-based methods and training-free models, verifying the benefit of carefully integrating tutorials. 
\ours{} also outperforms AgentTrek trained with synthetic trajectories converted from tutorials. 
We hypothesize that AgentTrek is hampered by dataset quality issues and still relies on high-quality trajectories to achieve strong performance.

\paragraph{Experiment on Online Tasks}  We evaluate \ours{} on AndroidWorld, an online environment requiring multi-step reasoning that mirrors real-world scenarios.
Remarkably, incorporating guided summarization leads to more than $10\%$ success rates compared to direct inference, which further validates the effectiveness of our proposed approach. 

\paragraph{Effect of Two-Stage Training}
The results also demonstrate that incorporating self-guided rejection sampling finetuning (RSF) consistently improves performance, emphasizing the benefit of enabling the model to self-evolve and avoid costly manual annotation.

\begin{table}[t]
\centering
\renewcommand\arraystretch{0.93}
\caption{The performance of \ours{} and frozen Qwen as textual guidance generators on the AndroidWorld benchmark.}
\label{tab:textual_guidance}
\resizebox{\linewidth}{!}{
\begin{tabular}{l|c}
\toprule
\textbf{Models} & \textbf{AW SR} \\
\midrule
Qwen2.5-VL-7B~\shortcite{bai2025qwen2} & 22.0 \\
~~~~~~~~~~~~~~~~~~ w/ vanilla RAG & 22.4 \\
~~~~~~~~~~~~~~~~~~ w/ textual guidance from frozen Qwen2.5-VL-7B & 27.5 \\
\rowcolor{lightcoral!20}\ours{}-7B & 35.3 \\
Qwen2.5-VL-72B~\shortcite{bai2025qwen2} & 35.0 \\
~~~~~~~~~~~~~~~~~~ w/ vanilla RAG & 37.5 \\
~~~~~~~~~~~~~~~~~~ w/ textual guidance from frozen Qwen2.5-VL-72B & 37.9 \\
\rowcolor{teal!12}\ours{}-72B & 45.7 \\
\bottomrule
\end{tabular}
}
\end{table}

\paragraph{Effect of Textual Guidance Generator} In order to evaluate the capability of \ours{} to generate high-quality textual guidance, we conduct additional experiments that leverage frozen Qwen2.5-VL-7B/72B as textual guidance generator for the Android World dataset. As shown in Table~\ref{tab:textual_guidance}, \ours{} consistently outperforms the frozen models generating textual guidance for both 7B and 72B backbones. This demonstrates that the task-adaptive fine-tuning and guidance optimization in \ours{} lead to substantial improvements over simply using an online LLM to generate guidance, validating the effectiveness of our approach.

\begin{figure}[t]
	\centering
     \subfigure[Different $k$]{
		\includegraphics[width=0.46\linewidth]{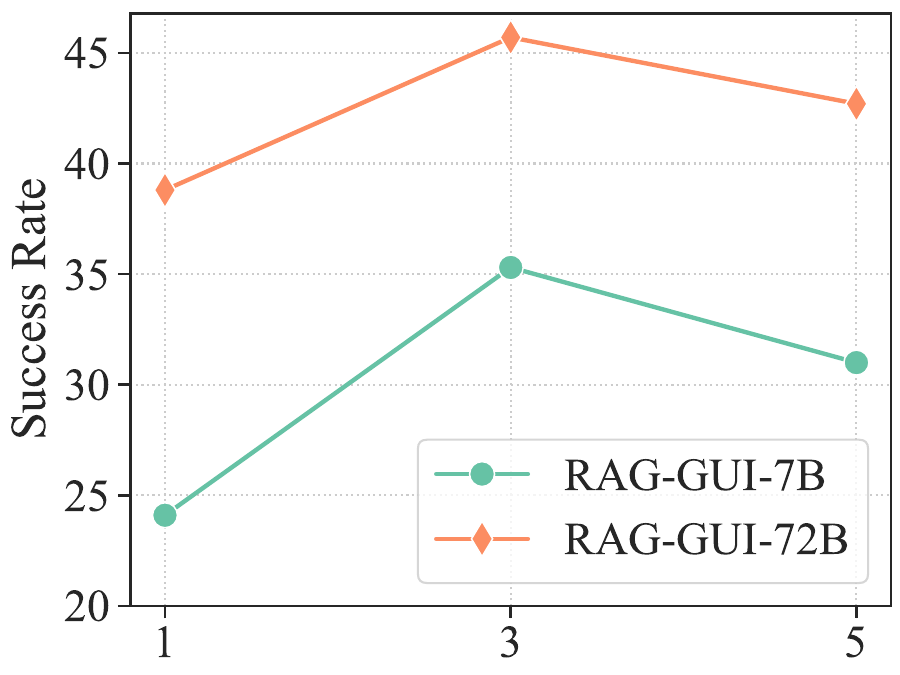}
		\label{fig:diff_k}
	}
     \subfigure[Alternative Designs]{
		\includegraphics[width=0.46\linewidth]{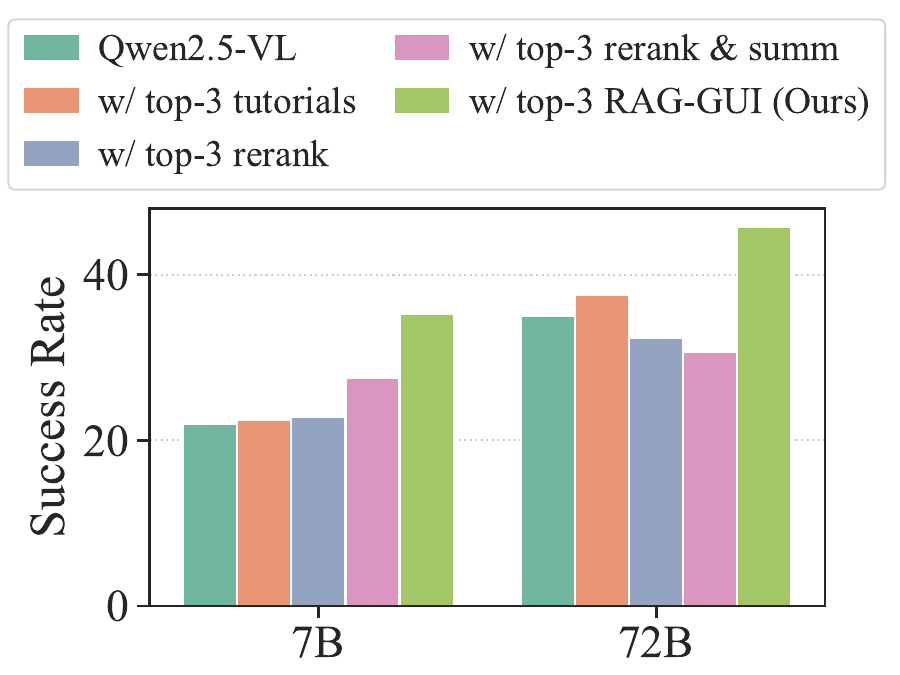}
		\label{fig:other_design}
	} 
\vspace{-1ex}
    \caption{Additional studies.\vspace{-1ex}}
\label{fig:add_study}
\end{figure}
\paragraph{Additional Studies}
We present additional analyses on guideline generation in Figure~\ref{fig:add_study}. As illustrated in Figure~\ref{fig:diff_k}, retrieving $k=3$ tutorials yields optimal performance—too many tutorials result in lengthy contexts that may confuse the model, while too few ($k=1$) risk missing relevant information. 
Furthermore, Figure~\ref{fig:other_design} compares our guideline generation framework against alternative designs, including prompting Qwen-2.5-VL-72B for reranking or using it solely as a summarization model without task-specific adaptation. The inferior performance of these variants verifies the effectiveness of our proposed guideline generation approach.
More cases studies are presented in Appendix \ref{app:case_study} for better illustration.

\section{Conclusion}

In this work, we introduce \ours{}, a VLM–based guideline generation framework that enables agents to effectively harness the vast information contained in web tutorials. We curate a large dataset of 2.6 million tutorials and adopt a two-stage training approach to produce high-quality guideline. Experiments across three benchmarks with two VLM backbones show consistent performance gains of \ours{}, ranging from 2.6\% to 13.3\% across model sizes. Notably, the improvements are more pronounced in online environments that closely simulate real-world scenarios, demonstrating the strong generalization capability of \ours{} in assisting VLM-based Agents.

% Footnotes are inserted with the \verb|\footnote| command.\footnote{This is a footnote.}

\section*{Limitations}
Our study is limited to inference-time evaluation without any model finetuning, which may constrain the achievable performance gains through adaptation. Furthermore, all experiments are performed exclusively on the Qwen-VL-series models, limiting the generalizability of our results to other architectures such as LLaVA~\citep{llava}. While incorporating guideline generation introduces additional inference latency, the consistent performance improvements justify this overhead. Nonetheless, future work could focus on optimizing the efficiency of both retrieval and guideline generation.

% \section*{Acknowledgments}

\bibliography{custom}

\begin{thebibliography}{42}
\providecommand{\natexlab}[1]{#1}

\bibitem[{Achiam et~al.(2023)Achiam, Adler, Agarwal, Ahmad, Akkaya, Aleman, Almeida, Altenschmidt, Altman, Anadkat et~al.}]{achiam2023gpt}
Josh Achiam, Steven Adler, Sandhini Agarwal, Lama Ahmad, Ilge Akkaya, Florencia~Leoni Aleman, Diogo Almeida, Janko Altenschmidt, Sam Altman, Shyamal Anadkat, and 1 others. 2023.
\newblock Gpt-4 technical report.
\newblock \emph{arXiv preprint arXiv:2303.08774}.

\bibitem[{Aksitov et~al.(2023)Aksitov, Miryoosefi, Li, Li, Babayan, Kopparapu, Fisher, Guo, Prakash, Srinivasan et~al.}]{aksitov2023rest}
Renat Aksitov, Sobhan Miryoosefi, Zonglin Li, Daliang Li, Sheila Babayan, Kavya Kopparapu, Zachary Fisher, Ruiqi Guo, Sushant Prakash, Pranesh Srinivasan, and 1 others. 2023.
\newblock Rest meets react: Self-improvement for multi-step reasoning llm agent.
\newblock \emph{arXiv preprint arXiv:2312.10003}.

\bibitem[{Anthropic(2025)}]{claude}
Anthropic. 2025.
\newblock \href {https://www.anthropic.com/news/3-5-models-and-computer-use} {Introducing computer use, a new claude 3.5 sonnet, and claude 3.5 haiku}.

\bibitem[{Awadalla et~al.(2024)Awadalla, Xue, Lo, Shu, Lee, Guha, Shen, Awadalla, Savarese, Xiong et~al.}]{awadalla2024mint}
Anas Awadalla, Le~Xue, Oscar Lo, Manli Shu, Hannah Lee, Etash Guha, Sheng Shen, Mohamed Awadalla, Silvio Savarese, Caiming Xiong, and 1 others. 2024.
\newblock Mint-1t: Scaling open-source multimodal data by 10x: A multimodal dataset with one trillion tokens.
\newblock \emph{Advances in Neural Information Processing Systems}, 37:36805--36828.

\bibitem[{Bai et~al.(2025)Bai, Chen, Liu, Wang, Ge, Song, Dang, Wang, Wang, Tang et~al.}]{bai2025qwen2}
Shuai Bai, Keqin Chen, Xuejing Liu, Jialin Wang, Wenbin Ge, Sibo Song, Kai Dang, Peng Wang, Shijie Wang, Jun Tang, and 1 others. 2025.
\newblock Qwen2. 5-vl technical report.
\newblock \emph{arXiv preprint arXiv:2502.13923}.

\bibitem[{Chai et~al.(2024)Chai, Huang, Niu, Xiao, Liu, Zhang, Gao, Ren, and Li}]{chai2024amex}
Yuxiang Chai, Siyuan Huang, Yazhe Niu, Han Xiao, Liang Liu, Dingyu Zhang, Peng Gao, Shuai Ren, and Hongsheng Li. 2024.
\newblock Amex: Android multi-annotation expo dataset for mobile gui agents.
\newblock \emph{arXiv preprint arXiv:2407.17490}.

\bibitem[{Chen et~al.(2024)Chen, Cui, Hu, Qin, Fang, Zhao, Wang, Liu, Chen, Huo et~al.}]{chen2024guicourse}
Wentong Chen, Junbo Cui, Jinyi Hu, Yujia Qin, Junjie Fang, Yue Zhao, Chongyi Wang, Jun Liu, Guirong Chen, Yupeng Huo, and 1 others. 2024.
\newblock Guicourse: From general vision language models to versatile gui agents.
\newblock \emph{arXiv preprint arXiv:2406.11317}.

\bibitem[{Cheng et~al.(2024{\natexlab{a}})Cheng, Sun, Chu, Xu, Li, Zhang, and Wu}]{cheng2024seeclick}
Kanzhi Cheng, Qiushi Sun, Yougang Chu, Fangzhi Xu, Yantao Li, Jianbing Zhang, and Zhiyong Wu. 2024{\natexlab{a}}.
\newblock Seeclick: Harnessing gui grounding for advanced visual gui agents.
\newblock \emph{arXiv preprint arXiv:2401.10935}.

\bibitem[{Cheng et~al.(2024{\natexlab{b}})Cheng, Sun, Chu, Xu, YanTao, Zhang, and Wu}]{cheng-etal-2024-seeclick}
Kanzhi Cheng, Qiushi Sun, Yougang Chu, Fangzhi Xu, Li~YanTao, Jianbing Zhang, and Zhiyong Wu. 2024{\natexlab{b}}.
\newblock \href {https://doi.org/10.18653/v1/2024.acl-long.505} {{S}ee{C}lick: Harnessing {GUI} grounding for advanced visual {GUI} agents}.
\newblock In \emph{Proceedings of the 62nd Annual Meeting of the Association for Computational Linguistics (Volume 1: Long Papers)}, pages 9313--9332, Bangkok, Thailand. Association for Computational Linguistics.

\bibitem[{Deng et~al.(2023)Deng, Gu, Zheng, Chen, Stevens, Wang, Sun, and Su}]{deng2023mind2web}
Xiang Deng, Yu~Gu, Boyuan Zheng, Shijie Chen, Sam Stevens, Boshi Wang, Huan Sun, and Yu~Su. 2023.
\newblock Mind2web: Towards a generalist agent for the web.
\newblock \emph{Advances in Neural Information Processing Systems}, 36:28091--28114.

\bibitem[{Gou et~al.(2025)Gou, Wang, Zheng, Xie, Chang, Shu, Sun, and Su}]{gou2025navigating}
Boyu Gou, Ruohan Wang, Boyuan Zheng, Yanan Xie, Cheng Chang, Yiheng Shu, Huan Sun, and Yu~Su. 2025.
\newblock \href {https://openreview.net/forum?id=kxnoqaisCT} {Navigating the digital world as humans do: Universal visual grounding for {GUI} agents}.
\newblock In \emph{The Thirteenth International Conference on Learning Representations}.

\bibitem[{Gur et~al.(2023)Gur, Furuta, Huang, Safdari, Matsuo, Eck, and Faust}]{gur2023real}
Izzeddin Gur, Hiroki Furuta, Austin Huang, Mustafa Safdari, Yutaka Matsuo, Douglas Eck, and Aleksandra Faust. 2023.
\newblock A real-world webagent with planning, long context understanding, and program synthesis.
\newblock \emph{arXiv preprint arXiv:2307.12856}.

\bibitem[{He et~al.(2024)He, Yao, Ma, Yu, Dai, Zhang, Lan, and Yu}]{he-etal-2024-webvoyager}
Hongliang He, Wenlin Yao, Kaixin Ma, Wenhao Yu, Yong Dai, Hongming Zhang, Zhenzhong Lan, and Dong Yu. 2024.
\newblock \href {https://doi.org/10.18653/v1/2024.acl-long.371} {{W}eb{V}oyager: Building an end-to-end web agent with large multimodal models}.
\newblock In \emph{Proceedings of the 62nd Annual Meeting of the Association for Computational Linguistics (Volume 1: Long Papers)}, pages 6864--6890, Bangkok, Thailand. Association for Computational Linguistics.

\bibitem[{Hong et~al.(2024)Hong, Wang, Lv, Xu, Yu, Ji, Wang, Wang, Dong, Ding et~al.}]{hong2024cogagent}
Wenyi Hong, Weihan Wang, Qingsong Lv, Jiazheng Xu, Wenmeng Yu, Junhui Ji, Yan Wang, Zihan Wang, Yuxiao Dong, Ming Ding, and 1 others. 2024.
\newblock Cogagent: A visual language model for gui agents.
\newblock In \emph{Proceedings of the IEEE/CVF Conference on Computer Vision and Pattern Recognition}, pages 14281--14290.

\bibitem[{Hurst et~al.(2024)Hurst, Lerer, Goucher, Perelman, Ramesh, Clark, Ostrow, Welihinda, Hayes, Radford et~al.}]{hurst2024gpt}
Aaron Hurst, Adam Lerer, Adam~P Goucher, Adam Perelman, Aditya Ramesh, Aidan Clark, AJ~Ostrow, Akila Welihinda, Alan Hayes, Alec Radford, and 1 others. 2024.
\newblock Gpt-4o system card.
\newblock \emph{arXiv preprint arXiv:2410.21276}.

\bibitem[{Joulin et~al.(2016)Joulin, Grave, Bojanowski, Douze, J{\'e}gou, and Mikolov}]{joulin2016fasttext}
Armand Joulin, Edouard Grave, Piotr Bojanowski, Matthijs Douze, H{\'e}rve J{\'e}gou, and Tomas Mikolov. 2016.
\newblock Fasttext. zip: Compressing text classification models.
\newblock \emph{arXiv preprint arXiv:1612.03651}.

\bibitem[{Lewis et~al.(2020)Lewis, Perez, Piktus, Petroni, Karpukhin, Goyal, K{\"u}ttler, Lewis, Yih, Rockt{\"a}schel et~al.}]{lewis2020retrieval}
Patrick Lewis, Ethan Perez, Aleksandra Piktus, Fabio Petroni, Vladimir Karpukhin, Naman Goyal, Heinrich K{\"u}ttler, Mike Lewis, Wen-tau Yih, Tim Rockt{\"a}schel, and 1 others. 2020.
\newblock Retrieval-augmented generation for knowledge-intensive nlp tasks.
\newblock \emph{Advances in neural information processing systems}, 33:9459--9474.

\bibitem[{Li et~al.(2024{\natexlab{a}})Li, Chen, Wang, Wang, Ye, Jin, Chen, He, Gao, Cui et~al.}]{li2024omnicorpus}
Qingyun Li, Zhe Chen, Weiyun Wang, Wenhai Wang, Shenglong Ye, Zhenjiang Jin, Guanzhou Chen, Yinan He, Zhangwei Gao, Erfei Cui, and 1 others. 2024{\natexlab{a}}.
\newblock Omnicorpus: A unified multimodal corpus of 10 billion-level images interleaved with text.
\newblock \emph{arXiv preprint arXiv:2406.08418}.

\bibitem[{Li et~al.(2024{\natexlab{b}})Li, Bishop, Li, Rawles, Campbell-Ajala, Tyamagundlu, and Riva}]{li2024effects}
Wei Li, William~E Bishop, Alice Li, Christopher Rawles, Folawiyo Campbell-Ajala, Divya Tyamagundlu, and Oriana Riva. 2024{\natexlab{b}}.
\newblock On the effects of data scale on ui control agents.
\newblock \emph{Advances in Neural Information Processing Systems}, 37:92130--92154.

\bibitem[{Liu et~al.(2023)Liu, Li, Wu, and Lee}]{llava}
Haotian Liu, Chunyuan Li, Qingyang Wu, and Yong~Jae Lee. 2023.
\newblock Visual instruction tuning.
\newblock \emph{Advances in neural information processing systems}, 36:34892--34916.

\bibitem[{Liu et~al.(2025)Liu, Jiang, Wang, Xu, Yu, Zhang, Zhao, and Wang}]{liu2025roserag}
Tianci Liu, Haoxiang Jiang, Tianze Wang, Ran Xu, Yue Yu, Linjun Zhang, Tuo Zhao, and Haoyu Wang. 2025.
\newblock Roserag: Robust retrieval-augmented generation with small-scale llms via margin-aware preference optimization.
\newblock \emph{arXiv preprint arXiv:2502.10993}.

\bibitem[{Lu et~al.(2024)Lu, Yang, Shen, and Awadallah}]{lu2024omniparser}
Yadong Lu, Jianwei Yang, Yelong Shen, and Ahmed Awadallah. 2024.
\newblock Omniparser for pure vision based gui agent.
\newblock \emph{arXiv preprint arXiv:2408.00203}.

\bibitem[{Ou et~al.(2024)Ou, Xu, Madaan, Liu, Lo, Sridhar, Sengupta, Roth, Neubig, and Zhou}]{ou2024synatra}
Tianyue Ou, Frank~F Xu, Aman Madaan, Jiarui Liu, Robert Lo, Abishek Sridhar, Sudipta Sengupta, Dan Roth, Graham Neubig, and Shuyan Zhou. 2024.
\newblock Synatra: Turning indirect knowledge into direct demonstrations for digital agents at scale.
\newblock \emph{arXiv preprint arXiv:2409.15637}.

\bibitem[{Qin et~al.(2025)Qin, Ye, Fang, Wang, Liang, Tian, Zhang, Li, Li, Huang et~al.}]{qin2025ui}
Yujia Qin, Yining Ye, Junjie Fang, Haoming Wang, Shihao Liang, Shizuo Tian, Junda Zhang, Jiahao Li, Yunxin Li, Shijue Huang, and 1 others. 2025.
\newblock Ui-tars: Pioneering automated gui interaction with native agents.
\newblock \emph{arXiv preprint arXiv:2501.12326}.

\bibitem[{Qwen et~al.(2025)Qwen, :, Yang, Yang, Zhang, Hui, Zheng, Yu, Li, Liu, Huang, Wei, Lin, Yang, Tu, Zhang, Yang, Yang, Zhou, Lin, Dang, Lu, Bao, Yang, Yu, Li, Xue, Zhang, Zhu, Men, Lin, Li, Tang, Xia, Ren, Ren, Fan, Su, Zhang, Wan, Liu, Cui, Zhang, and Qiu}]{qwen2025qwen25technicalreport}
Qwen, :, An~Yang, Baosong Yang, Beichen Zhang, Binyuan Hui, Bo~Zheng, Bowen Yu, Chengyuan Li, Dayiheng Liu, Fei Huang, Haoran Wei, Huan Lin, Jian Yang, Jianhong Tu, Jianwei Zhang, Jianxin Yang, Jiaxi Yang, Jingren Zhou, and 25 others. 2025.
\newblock \href {https://arxiv.org/abs/2412.15115} {Qwen2.5 technical report}.
\newblock \emph{Preprint}, arXiv:2412.15115.

\bibitem[{Rawles et~al.(2024)Rawles, Clinckemaillie, Chang, Waltz, Lau, Fair, Li, Bishop, Li, Campbell-Ajala et~al.}]{rawles2024androidworld}
Christopher Rawles, Sarah Clinckemaillie, Yifan Chang, Jonathan Waltz, Gabrielle Lau, Marybeth Fair, Alice Li, William Bishop, Wei Li, Folawiyo Campbell-Ajala, and 1 others. 2024.
\newblock Androidworld: A dynamic benchmarking environment for autonomous agents.
\newblock \emph{arXiv preprint arXiv:2405.14573}.

\bibitem[{Rawles et~al.(2023)Rawles, Li, Rodriguez, Riva, and Lillicrap}]{rawles2023androidinthewild}
Christopher Rawles, Alice Li, Daniel Rodriguez, Oriana Riva, and Timothy Lillicrap. 2023.
\newblock Androidinthewild: A large-scale dataset for android device control.
\newblock \emph{Advances in Neural Information Processing Systems}, 36:59708--59728.

\bibitem[{Shi et~al.(2024)Shi, Min, Yasunaga, Seo, James, Lewis, Zettlemoyer, and Yih}]{shi-etal-2024-replug}
Weijia Shi, Sewon Min, Michihiro Yasunaga, Minjoon Seo, Richard James, Mike Lewis, Luke Zettlemoyer, and Wen-tau Yih. 2024.
\newblock \href {https://doi.org/10.18653/v1/2024.naacl-long.463} {{REPLUG}: Retrieval-augmented black-box language models}.
\newblock In \emph{Proceedings of the 2024 Conference of the North American Chapter of the Association for Computational Linguistics: Human Language Technologies (Volume 1: Long Papers)}, pages 8371--8384, Mexico City, Mexico. Association for Computational Linguistics.

\bibitem[{Sun et~al.(2024)Sun, Cheng, Ding, Jin, Wang, Xu, Wu, Jia, Chen, Liu et~al.}]{sun2024genesis}
Qiushi Sun, Kanzhi Cheng, Zichen Ding, Chuanyang Jin, Yian Wang, Fangzhi Xu, Zhenyu Wu, Chengyou Jia, Liheng Chen, Zhoumianze Liu, and 1 others. 2024.
\newblock Os-genesis: Automating gui agent trajectory construction via reverse task synthesis.
\newblock \emph{arXiv preprint arXiv:2412.19723}.

\bibitem[{Wang et~al.(2022)Wang, Yang, Huang, Jiao, Yang, Jiang, Majumder, and Wei}]{wang2022text}
Liang Wang, Nan Yang, Xiaolong Huang, Binxing Jiao, Linjun Yang, Daxin Jiang, Rangan Majumder, and Furu Wei. 2022.
\newblock Text embeddings by weakly-supervised contrastive pre-training.
\newblock \emph{arXiv preprint arXiv:2212.03533}.

\bibitem[{Wu et~al.(2025)Wu, Wu, Xu, Wang, Sun, Jia, Cheng, Ding, Chen, Liang, and Qiao}]{wu2025osatlas}
Zhiyong Wu, Zhenyu Wu, Fangzhi Xu, Yian Wang, Qiushi Sun, Chengyou Jia, Kanzhi Cheng, Zichen Ding, Liheng Chen, Paul~Pu Liang, and Yu~Qiao. 2025.
\newblock \href {https://openreview.net/forum?id=n9PDaFNi8t} {{OS}-{ATLAS}: Foundation action model for generalist {GUI} agents}.
\newblock In \emph{The Thirteenth International Conference on Learning Representations}.

\bibitem[{Xie et~al.(2024)Xie, Zhang, Chen, Li, Zhao, Cao, Hua, Cheng, Shin, Lei et~al.}]{xie2024osworld}
Tianbao Xie, Danyang Zhang, Jixuan Chen, Xiaochuan Li, Siheng Zhao, Ruisheng Cao, Toh~J Hua, Zhoujun Cheng, Dongchan Shin, Fangyu Lei, and 1 others. 2024.
\newblock Osworld: Benchmarking multimodal agents for open-ended tasks in real computer environments.
\newblock \emph{Advances in Neural Information Processing Systems}, 37:52040--52094.

\bibitem[{Xu et~al.(2025{\natexlab{a}})Xu, Liu, Nag, Dai, Xie, Tang, Luo, Li, Ho, Yang, and He}]{xu-etal-2025-simrag}
Ran Xu, Hui Liu, Sreyashi Nag, Zhenwei Dai, Yaochen Xie, Xianfeng Tang, Chen Luo, Yang Li, Joyce~C. Ho, Carl Yang, and Qi~He. 2025{\natexlab{a}}.
\newblock \href {https://doi.org/10.18653/v1/2025.naacl-long.575} {{S}im{RAG}: Self-improving retrieval-augmented generation for adapting large language models to specialized domains}.
\newblock In \emph{Proceedings of the 2025 Conference of the Nations of the Americas Chapter of the Association for Computational Linguistics: Human Language Technologies (Volume 1: Long Papers)}, pages 11534--11550, Albuquerque, New Mexico. Association for Computational Linguistics.

\bibitem[{Xu et~al.(2025{\natexlab{b}})Xu, Lu, Shen, Wang, Wang, Mao, Xiong, and Yu}]{xu2025agenttrek}
Yiheng Xu, Dunjie Lu, Zhennan Shen, Junli Wang, Zekun Wang, Yuchen Mao, Caiming Xiong, and Tao Yu. 2025{\natexlab{b}}.
\newblock \href {https://openreview.net/forum?id=EEgYUccwsV} {Agenttrek: Agent trajectory synthesis via guiding replay with web tutorials}.
\newblock In \emph{The Thirteenth International Conference on Learning Representations}.

\bibitem[{Xu et~al.(2024)Xu, Wang, Wang, Lu, Xie, Saha, Sahoo, Yu, and Xiong}]{xu2024aguvis}
Yiheng Xu, Zekun Wang, Junli Wang, Dunjie Lu, Tianbao Xie, Amrita Saha, Doyen Sahoo, Tao Yu, and Caiming Xiong. 2024.
\newblock Aguvis: Unified pure vision agents for autonomous gui interaction.
\newblock \emph{arXiv preprint arXiv:2412.04454}.

\bibitem[{Yan et~al.(2023)Yan, Yang, Zhu, Lin, Li, Wang, Yang, Zhong, McAuley, Gao et~al.}]{yan2023gpt}
An~Yan, Zhengyuan Yang, Wanrong Zhu, Kevin Lin, Linjie Li, Jianfeng Wang, Jianwei Yang, Yiwu Zhong, Julian McAuley, Jianfeng Gao, and 1 others. 2023.
\newblock Gpt-4v in wonderland: Large multimodal models for zero-shot smartphone gui navigation.
\newblock \emph{arXiv preprint arXiv:2311.07562}.

\bibitem[{Yu et~al.(2025)Yu, Tang, Xu, Cui, Ran, Yan, Liu, Wang, Han, Liu, and Sun}]{yu2025visrag}
Shi Yu, Chaoyue Tang, Bokai Xu, Junbo Cui, Junhao Ran, Yukun Yan, Zhenghao Liu, Shuo Wang, Xu~Han, Zhiyuan Liu, and Maosong Sun. 2025.
\newblock \href {https://openreview.net/forum?id=zG459X3Xge} {Vis{RAG}: Vision-based retrieval-augmented generation on multi-modality documents}.
\newblock In \emph{The Thirteenth International Conference on Learning Representations}.

\bibitem[{Yu et~al.(2024{\natexlab{a}})Yu, Zhang, Pan, Cao, Ma, Li, Wang, and Yu}]{yu-etal-2024-chain}
Wenhao Yu, Hongming Zhang, Xiaoman Pan, Peixin Cao, Kaixin Ma, Jian Li, Hongwei Wang, and Dong Yu. 2024{\natexlab{a}}.
\newblock \href {https://doi.org/10.18653/v1/2024.emnlp-main.813} {Chain-of-note: Enhancing robustness in retrieval-augmented language models}.
\newblock In \emph{Proceedings of the 2024 Conference on Empirical Methods in Natural Language Processing}, pages 14672--14685, Miami, Florida, USA. Association for Computational Linguistics.

\bibitem[{Yu et~al.(2024{\natexlab{b}})Yu, Ping, Liu, Wang, You, Zhang, Shoeybi, and Catanzaro}]{yu2024rankrag}
Yue Yu, Wei Ping, Zihan Liu, Boxin Wang, Jiaxuan You, Chao Zhang, Mohammad Shoeybi, and Bryan Catanzaro. 2024{\natexlab{b}}.
\newblock \href {https://openreview.net/forum?id=S1fc92uemC} {Rank{RAG}: Unifying context ranking with retrieval-augmented generation in {LLM}s}.
\newblock In \emph{The Thirty-eighth Annual Conference on Neural Information Processing Systems}.

\bibitem[{Zhang et~al.(2025)Zhang, Shang, Gao, Zhang, Xie, Ma, Yuan, Wu, Zhu, and Li}]{zhang2025tongui}
Bofei Zhang, Zirui Shang, Zhi Gao, Wang Zhang, Rui Xie, Xiaojian Ma, Tao Yuan, Xinxiao Wu, Song-Chun Zhu, and Qing Li. 2025.
\newblock Tongui: Building generalized gui agents by learning from multimodal web tutorials.
\newblock \emph{arXiv preprint arXiv:2504.12679}.

\bibitem[{Zheng et~al.(2024)Zheng, Gou, Kil, Sun, and Su}]{zheng2024gptvision}
Boyuan Zheng, Boyu Gou, Jihyung Kil, Huan Sun, and Yu~Su. 2024.
\newblock \href {https://openreview.net/forum?id=piecKJ2DlB} {{GPT}-4v(ision) is a generalist web agent, if grounded}.
\newblock In \emph{Forty-first International Conference on Machine Learning}.

\bibitem[{Zhou et~al.(2024)Zhou, Xu, Zhu, Zhou, Lo, Sridhar, Cheng, Ou, Bisk, Fried, Alon, and Neubig}]{zhou2024webarena}
Shuyan Zhou, Frank~F. Xu, Hao Zhu, Xuhui Zhou, Robert Lo, Abishek Sridhar, Xianyi Cheng, Tianyue Ou, Yonatan Bisk, Daniel Fried, Uri Alon, and Graham Neubig. 2024.
\newblock \href {https://openreview.net/forum?id=oKn9c6ytLx} {Webarena: A realistic web environment for building autonomous agents}.
\newblock In \emph{The Twelfth International Conference on Learning Representations}.

\end{thebibliography}
\clearpage
\appendix

\section{Details for Evaluation Tasks}
\label{app:datasets}
Android World assesses performance within a virtual Android emulator, with 116 distinct tasks across 20 mobile applications. For this dataset, we report Step accuracy, which measures the correctness of the final stage of task execution.
Multimodal-Mind2Web focuses on interactions within web environments, with 1,013 tasks spanning 100 different websites. We utilize three metrics for evaluation: Element accuracy (Ele. Acc.), Operation F1 (Op. F1), and Step success rate (Step SR).
Android Control is designed for the mobile environment. Following the methodology of \citet{li2024effects}, we randomly sample 500 tasks to form our test set. Performance on this dataset is measured by Step accuracy. Notably, Android Control includes both high-level instructions, requiring the model to simultaneously plan and execute actions, and low-level instructions, where the model only needs to execute a predefined action.
\section{Details for Tutorial Collection}
\label{app:tutorial}
\paragraph{FastText Filtering} Given the substantial scale of the MINT (1054M documents) and OmniCorpus (988M documents) datasets, we employ the computationally efficient FastText classifier~\citep{joulin2016fasttext} for an initial filtering of potential tutorials. Leveraging the inherent feature of the WikiHow corpus, which categorizes its articles by topic, we curated a positive training set by extracting approximately 24K samples classified under the "Computers and Electronics" category. To create a balanced training set, we randomly selected 12K negative samples from WikiHow articles belonging to other categories and an additional 12K random samples from the MINT dataset. This resulted in a training corpus of 48K labeled examples. The trained FastText classifier is then applied to filter the MINT and OmniCorpus datasets, after which 26.5M documents from MINT and 52M documents from OmniCorpus are retained, forming a preliminary candidate set of tutorials.

\paragraph{Deduplication}
To mitigate redundancy within the candidate tutorial set derived from MINT and OmniCorpus (e.g., content duplication across different URLs and overlap between the two datasets), we performed content-based deduplication on the filtered documents from the previous stage. This process yields a refined set of approximately 7.5M unique documents from MINT and 24M unique documents from OmniCorpus.

\paragraph{LLM Labeling}
To achieve a more precise selection of GUI-related tutorials, we employ Qwen2.5-7b-it~\citep{qwen2025qwen25technicalreport} for a subsequent classification step. We prompt the LLM to analyze the content of each document and identify those specifically providing instructions for GUI tasks. This more granular filtering process further reduces the number of false positives, resulting in a high-quality set of approximately 0.74M GUI tutorials from MINT and 1.8M from OmniCorpus. The prompt format is detailed in Appendix~\ref{apd:promp_llm_label}. Aggregating these with the initial 24K positive samples sourced from WikiHow yields a final pool of 2.6M GUI tutorials, which serves as the task-adaptive guidance resource for the subsequent experiments.

\section{Prompt Format for LLM Labeling}
\label{apd:promp_llm_label}
Figure \ref{fig:web_tutorial_labels} shows the prompts used for estimating whether the content is a GUI-related tutorial.
\begin{figure}[htbp]
\centering
\begin{tcolorbox}[
    colback=gray!15,
    colframe=gray!75,
    % title=Rationale Generation on HotPotQA and 2WikiMultiHopQA,
    fonttitle=\large\bfseries\sffamily\color{white},
    coltitle=white,
    bottomrule=0pt,
    toprule=0pt,
    leftrule=0pt,
    rightrule=0pt,
    rounded corners,
    % width=0.9\linewidth
]
\textbf{System Prompt:} You are an assistant that classifies content based on specific criteria. Your task is to evaluate whether a given piece of content serves as a tutorial specifically related to graphical user interfaces (GUI), such as for web applications, desktop applications, or operating systems.

\# Classification Criteria
The content qualifies as a GUI-related tutorial if it meets the following conditions:
1. It includes a task description outlining what needs to be achieved.
2. It provides clear step-by-step instructions for interacting with a GUI, such as:
- Step 1: Open the application
- Step 2: Navigate to the settings menu.

\medskip
\textbf{User Prompt:} Given the below content, predict if the content is a GUI-related tutorial or not. Output 'Yes' if the above content is a GUI-related tutorial and 'No' if it is not. Provide only 'Yes' or 'No' as the output.

\{content\}

\medskip
\textbf{Assistant Prompt:} \\
\textbf{Output:} \{Generated relevant label\}
\end{tcolorbox}
\caption{Prompt for using Qwen-2.5 for generating quality labels for web tutorials.}
\label{fig:web_tutorial_labels}
\end{figure}
\section{Prompt Format for Inference}
The prompt template for guidance generation is listed in Figure \ref{fig:guidance_gen}.
\begin{figure}[htbp]
\centering
\begin{tcolorbox}[
    colback=gray!15,
    colframe=gray!75,
    % title=Rationale Generation on HotPotQA and 2WikiMultiHopQA,
    fonttitle=\large\bfseries\sffamily\color{white},
    coltitle=white,
    bottomrule=0pt,
    toprule=0pt,
    leftrule=0pt,
    rightrule=0pt,
    rounded corners,
    % width=0.9\linewidth
]
\textbf{System Prompt:} You are a helpful assistant that aim to use a tutorial for completing a specific GUI-based task. Given a query, previous actions and a related tutorial, your task is to first identify the relevance between the tutorial and the task and previous actions. Then, if the tutorial is relevant, please generate a concise summary for the tutorial with the following requirements:
- 1. You should only retain the most relevant information from the tutorial that help to solve the task conditioned on previous actions.
- 2.Please make sure to include detailed guidance from the tutorial if it is helpful to solve the problem based on the current state.
- 3. If the tutorial is not helpful or relevant to the task, then please only generate the score without summary.
Use the following format in your output:
<score>
[the relevance score (0 or 1)]
</score>
<summary>
[Your summary of the tutorial conditioned on the task and previous actions]
</summary>

\medskip
\textbf{User Prompt:} The user query: \{instruction\} \\
Task progress (You have done the following operation on the current device): \{previous\_actions\}; \\
Tutorial: \{tutorial\} 

\medskip
\textbf{Assistant Prompt:} \\
\textbf{Output:} \{Model generated guidance\}
\end{tcolorbox}
\caption{Prompt for guidance generation.}
\label{fig:guidance_gen}
\end{figure}

\section{Case Study}
\label{app:case_study}
\begin{table*}[t]
\centering
\caption{A case study from Android Control demonstrating the task guidance generated by \ours{}.
\textcolor{green}{Green text} stands for the alignment of guidance and action goal while \textcolor{blue}{blue text} justify the ability of \ours{} in generating customized guidance by taking the previous actions into the consideration. 
}
\label{tab:case_study}
\resizebox{\linewidth}{!}{
    \begin{tabular}{p{2.5cm}p{22.5cm}}
       \toprule
       \bf Goal & I want to search for a flight from \textcolor{green}{Rotterdam} to \textcolor{green}{Puerto Natales} to visit my favorite travel destination Patagonia on the MakeMyTrip app for \textcolor{green}{11th January}. \\
       \midrule 
       \bf Previous \newline Actions & Step 1: Open MakeMyTrip app.\newline Step 2: \textcolor{blue}{Click on the Flights option.} \\
       \midrule
        \bf Retrieved \newline Tutorial & Search\newline In the Main Page select the ``Flights'' tab, and a form is shown with the information you submit for search flights.\newline Complete the Form \newline In the form, enter the information for your reservation: Depart and return cities, depart and return dates, amount of tickets, etc. \newline After filling out the form, click on:\newline Select Flights\newline You will see a list with the flights that match your search criteria in previous step.\newline Verify pricing information, dates, etc; and select the flight you want by clicking on this button:\newline Verify\newline The reservation summary information is shown,with Flight data, prices, purchase conditions, etc.\newline To continue, click on the following button:\newline Enter your Data\newline The form is shown to collect the information for passengers, contact, etc.\newline Complete the form and click on ``Reserve''.\newline Payment Information\newline Now the form to enter payment information, that could be Credit Card or Deposit. IMPORTANT: Your credit card is not charged on this step. It is used only for reservation validation. A Sales executive will contact you to complete the payment process.\newline Click on ``Accept Terms and Conditions'' and then on ``Confirm Trip''.\newline Reservation Complete\newline Congratulations! you have reserved your flight ticket. A Sales Executive will contact you to complete the payment and issue the Electronic Ticket.\newline Packages\newline Hotels\newline Air Tickets\newline Search for Tourism Packages\newline Ingrese el nombre del hotel o la ciudad:\newline Partners Hotels recommended \\
        \midrule
        \bf Task Guidance Generated by \ours{} & In the main page of the MakeMyTrip app, \textcolor{blue}{after selecting the ``Flights'' tab}, proceed to fill out the flight search form by entering the departure city as \textcolor{green}{Rotterdam}, the destination city as \textcolor{green}{Puerto Natales}, and set the travel date to \textcolor{green}{11th January}. After completing the form, initiate the search to view available flights matching these criteria. Select the preferred flight from the search results to verify details such as pricing and dates. Continue by entering passenger and contact information, then proceed to payment information where you accept terms and confirm the trip to complete the booking. \\
       \bottomrule
    \end{tabular}
%\end{center}
}
\end{table*}

We present a case study in Table~\ref{tab:case_study} from Android Control to demonstrate the task guidance generated by \ours{} based on goal, previous actions and the retrieved tutorial. We observe that the retrieved tutorial contains noise, much of which is irrelevant to the specific task at hand.. However, the task guidance generated by \ours{} is much more concise, while still retaining the important information. Furthermore, the generated guidance seamlessly integrates contextual data derived from both the stated goal and the prior actions. Crucially, it provides specific, actionable cues on how to execute the subsequent step, directly addressing the requirements of the current task in light of the performed interactions.

\end{document}